\documentclass[letterpaper, 10 pt, conference]{ieeeconf}  

\IEEEoverridecommandlockouts                              
\usepackage{graphicx} 

\usepackage[margin=0pt,font=small,labelfont=bf]{caption}

\usepackage{amsmath,amssymb,amsfonts}
\usepackage{bm}

\usepackage{pifont}
\usepackage{stfloats}
\usepackage{subcaption}
\usepackage{lipsum}  

\usepackage{cite}

\usepackage{tabularx}
\usepackage{algorithmic}

\usepackage{booktabs}
\usepackage{multirow}
\usepackage{booktabs} 
\usepackage[colorlinks,bookmarksopen,bookmarksnumbered,linkcolor=blue,citecolor=blue,urlcolor=blue]{hyperref}

\usepackage[a-1b]{pdfx}

\title{\LARGE \bf Dynamic Modeling and Analysis of Impact-resilient MAVs Undergoing \\High-speed and Large-angle Collisions with the Environment
}

\author{Zhichao~Liu and Konstantinos~Karydis
\thanks{The authors are with the Dept. of Electrical and Computer Engineering, University of California, Riverside. Email: \{zliu157, karydis\}@ucr.edu. We gratefully acknowledge the support of NSF \#  IIS-1910087 and \# IIS-1901379, ARL \# W911NF-18-1-0266, and ONR  \# N00014-19-1-2264. Any opinions, findings, and conclusions or recommendations expressed in this material are those of the authors and do not necessarily reflect the views of the funding agencies.
}
}

\begin{document}

\maketitle
\thispagestyle{empty}
\pagestyle{empty}

\begin{abstract}

Micro Aerial Vehicles (MAVs) often face a high risk of collision during autonomous flight, particularly in cluttered and unstructured environments. To mitigate the collision impact on sensitive onboard devices, resilient MAVs with mechanical protective cages and reinforced frames are commonly used. However, compliant and impact-resilient MAVs offer a promising alternative by reducing the potential damage caused by impacts. In this study, we present novel findings on the impact-resilient capabilities of MAVs equipped with passive springs in their compliant arms. We analyze the effect of compliance through dynamic modeling and demonstrate that the inclusion of passive springs enhances impact resilience. The impact resilience is extensively tested to stabilize the MAV following wall collisions under high-speed and large-angle conditions. Additionally, we provide comprehensive comparisons with rigid MAVs to better determine the tradeoffs in flight by embedding compliance onto the robot's frame.

\end{abstract}

\section{INTRODUCTION}

Micro Aerial Vehicles (MAVs) have the potential to function as versatile platforms for sensor-based exploration and navigation. In recent years, there has been an increasing interest in deploying MAVs in challenging environments, including confined areas~\cite{tranzatto2021cerberus, de2021resilient, lew2019contact} and cluttered spaces~\cite{zhou2021raptor, mulgaonkar2017robust}. Autonomous missions in these complex environments, particularly at high speeds, significantly increase the risk of collisions.  To mitigate the risk, multiple impact-resilient vehicles~\cite{de2021resilient, Patnaik, zha2020collision, liu2021arq} have been developed, with applications to collision-inclusive navigation~\cite{lu2019optimal, zlumotion2020, zha2021exploiting, zlu2021iros, lu2022online}. 

Mechanical protective cages and reinforced frames have been explored in an effort to protect hardware during crashes in MAVs~\cite{klaptocz2013euler, naldi2014robust, khedekar2019contact, chui2016dynamics, dicker2017quadrotor, dicker2018recovery, mulgaonkar2020tiercel}. However, in high-speed collisions, the impact energy is still directly transferred to the robot, potentially damaging sensitive electronic components such as the IMU, camera, or LiDAR, despite the use of a protective mechanical design~\cite{patnaik2021towards}. To address this challenge, novel MAV designs have been proposed that integrate compliance into the airframe to reduce the impact. Examples of compliant MAV designs include those with purely soft frames~\cite{mintchev2017insect, sareh2018rotorigami, shu2019quadrotor}, icosahedron tensegrity structures~\cite{zha2020collision}, external compliant flaps~\cite{de2021resilient}, and foldable arms with passive springs~\cite{Patnaik, liu2021arq, liu2023contact}. 
These works have demonstrated the efficacy of compliant MAVs in reducing effects of impact and helping survive collisions. 

However, several interesting questions regarding compliant MAVs remain open. These include impact modeling and the ability to stabilize from large-angle (namely pitch) collisions, as well as an in-depth analysis and comparison with rigid MAVs to evaluate the tradeoffs of embedding compliance into the airframe. 
Although modeling impacts has been thoroughly studied in the literature~\cite{gilardi2002literature}, only a few existing works on rigid aerial robots take it into consideration. Notable examples include the impulse-momentum principle model in~\cite{mulgaonkar2017robust} and viscoelastic models in~\cite{chui2016dynamics, dicker2018recovery}. Additionally, it is necessary to model the embedded compliance to describe its effect on impact reduction. Prior work~\cite{Patnaik} uses a damper and spring model to describe the arm length. However, this method assumes massless compliant arms that limit precise description of fast and periodic movements~\cite{cherouvim2009control}.

\begin{figure}[!t]
\vspace{4pt}
	\centering
	\includegraphics[trim={0cm 1.85cm 0cm 0.5cm}, clip, width=0.99\linewidth]{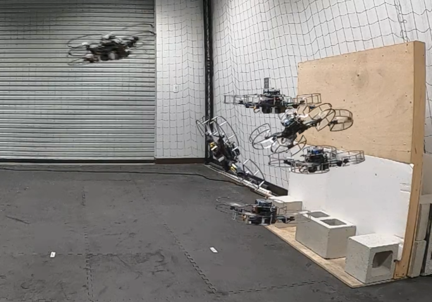}
	\caption{Our approach enables compliant MAVs to rapidly stabilize from collisions with walls at speed of up to $3.5$\;m/s and sustain post-impact flight by tracking aggressive recovery trajectories. A supplemental video demonstrates all experiments conducted in this work and it can be accessed at \url{https://youtu.be/b0xU2CzQWRg}.} 
	\label{fig:arq_35}
	\vspace{-12pt}
\end{figure}

In addition, the majority of existing works on MAV collision and recovery share the assumption that the contact (pitch) angle is close to zero. This restriction partly comes from linear flight controllers~\cite{de2021resilient}, however, fast autonomous flight of MAVs involves large accelerations and attitude changes~\cite{falanga2017aggressive, landry2016aggressive}. 
Therefore, the ability to stabilize large-angle collisions remains crucial for fast and autonomous missions of MAVs in challenging environments. Notable exceptions include related work on rigid aerial robots~\cite{dicker2017quadrotor, dicker2018recovery} that show successful collision recovery with contact angles up to $30^{\circ}$ in physical tests. 
Further, introducing compliance within the airframe can negatively affect the free flight performance by increasing weight and reducing flight time. Compliance may also introduce modeling errors and degrade trajectory tracking performance based on model-based control. Thus, it is important to study the free flight performance of compliant MAVs in comparison to rigid ones.


Different from our previous work~\cite{liu2021arq}, this work studies dynamic modeling to help understand the effect of the compliance. The impact resilience is reinforced to stabilize from both high-speed and large-angle collisions. Furthermore, we include comprehensive comparisons with a rigid MAV to validate the efficacy of our proposed compliant design and control methods. 
We introduce dynamics modeling for both compliant arms and contact. Physical collision experiments against both rigid walls and soft mats are used to verify the proposed models. Taking advantage of the compliant airframe and a geometric tracking controller, the robot can survive collisions against walls at speeds of up to $3.5$ m/s. The robot is also observed to survive collisions with contact angles of up to $45^\circ$ at a speed of $2$ m/s. Multiple repeated physical experiments demonstrate 100\% success rates for all collision tests ranging from high-speed to large-angle ones. 



\section{COMPLIANT MAV DESIGN OVERVIEW}\label{sec:design}

\begin{figure*}[!t]
\vspace{6pt}
	\centering
    \includegraphics[width=0.8\linewidth]{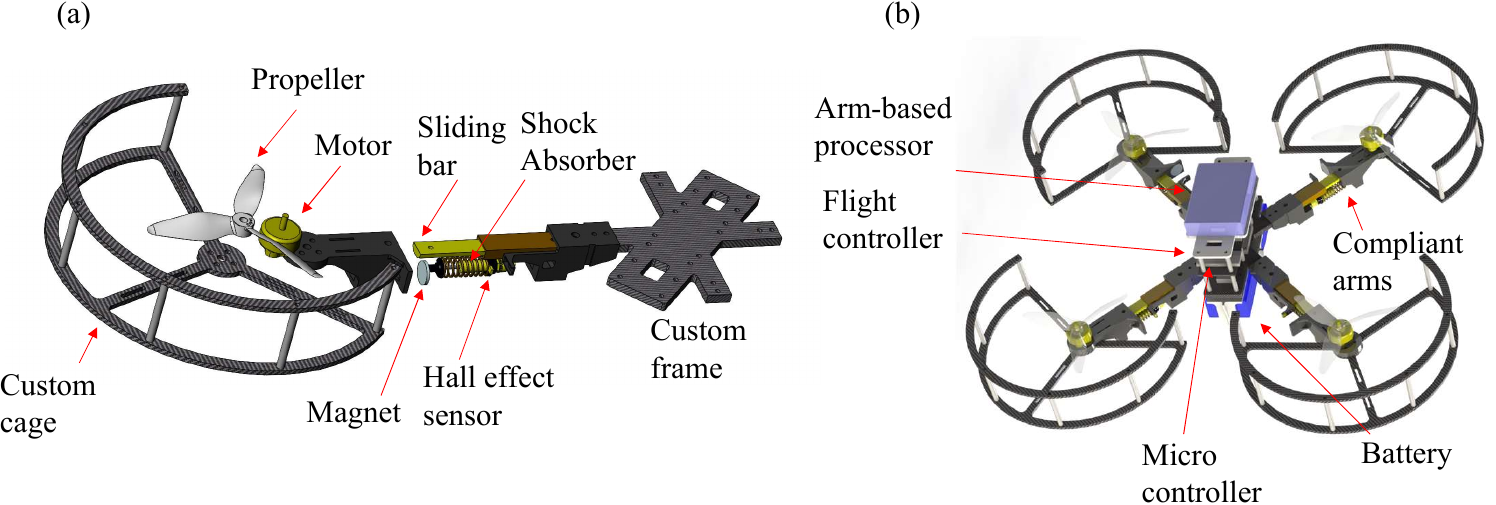}
	\caption{(a) The exploded view of the compliant arm design in the computer-aided design (CAD) software. (b) A CAD rendering image of the compliant MAV featuring the novel foldable arms to reduce impact.}
	\label{fig:design}
	\vspace{-18pt}
\end{figure*}



The section summarizes the hardware design of the resilient aerial robot (ARQ). This design shares similar principles as in our prior work~\cite{liu2021arq} but has some engineering modifications that help increase robustness.\footnote{~We include information regarding our MAV design to make this paper self-contained, and refer to~\cite{liu2021arq} for more details about the hardware.}

Figure~\ref{fig:design}a shows an exploded view of the compliant arm design. 
It features a prismatic joint, shock absorber, Hall effect sensor, magnet and custom protective cage. The prismatic joint is built based on a metallic sliding bar, while the shock absorber is directly taken from 1/18 radio-control cars. An A1302 ratiometric linear Hall effect sensor is fixed on the prismatic joint to measure the magnetic intensity. The adapters connecting the prismatic joint and shock absorber are 3D-printed (Markforged Mark II, onyx material with carbon fiber add-in). The custom cages and frames are fabricated in lightweight carbon fiber sheets (tensile strength 120,000-175,000 psi) using a Stepcraft D.600 CNC router with enclosure and milling bath. 


The overall platform (Fig.~\ref{fig:design}b) integrates an Arm-based multi-core processor (Odroid XU4 2Ghz) running high-level computing tasks. A flight controller (Pixhawk 4 Mini) is used for the autopilot (PX4). A microcontroller (Arduino Nano) is used for analog to digital conversion for Hall effect sensor data processing. The free length of compliant arms measures $0.19$\;m. Protective cages have a radius of $0.11$\;m. Thus, the robot measures $0.6$\;m from the cage tip to tip. 
 Key robot features are given in Table~\ref{tb:arq_values}. 

\begin{table}[htb]
\vspace{-4pt}
\renewcommand{\arraystretch}{1.3}
\caption{Key Features of Compliant and Rigid MAVs}
\label{tb:arq_values}
\centering
\begin{tabular}{l| c| c| c}
\toprule
\bfseries Descriptions & \bfseries  ARQ & \bfseries Quad  &\bfseries Units \\
\toprule
 Arm length ($L$)  & $0.19$ & $0.19$  & m\\
 \toprule
 Size (cage tip to tip) & $0.60$ & $0.60$ & m\\ 
\midrule
 Weight w/o battery  & $1.034$ & $0.844$& kg\\
 \midrule
 Moments of inertia ($I_{xx}$)  & $0.0100$ & $0.0092$ & $\text{kg} \cdot \text{m}^2$\\
 \midrule
 Moments of inertia ($I_{yy}$)  & $0.0116$ & $0.0107$  & $\text{kg} \cdot \text{m}^2$\\
 \midrule
 Moments of inertia ($I_{zz}$)  & $0.0197$ & $0.0179$ & $\text{kg} \cdot \text{m}^2$\\
  \midrule
 Maximum payload & $1.38$  & $1.64$ & kg\\
 \midrule
 Flight Time (Hover)  & $461$ & $573$ &sec\\
 \bottomrule
\end{tabular}
\vspace{-8pt}
\end{table}

\begin{figure}[!ht]
\vspace{-9pt}
	\centering
	\includegraphics[width=0.85\linewidth]{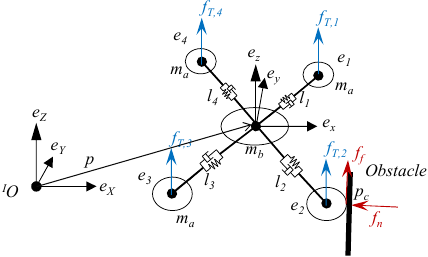}
	\caption{Inertial and body-fixed frames description. The robot is represented by five rigid bodies during impact. Thrust force (blue) and contact force (red) are shown in the body-fixed frame. Vector $e_j$ is the relative position of the arm $j$ in the body-fixed frame, while $p$ is the position of the main body in the inertial frame.}
	\label{fig:frame}
	\vspace{-12pt}
\end{figure}

\section{DYNAMIC MODELING}\label{sec:model}
Different from dynamics of rigid rotorcraft with elastic contacts~\cite{mueller2021design}, we study herein the embedded compliance into the MAV arms in a way similar to legged locomotion compliance modeling~\cite{vasilopoulos2014compliant}. We study the dynamics in free flight and collision scenarios separately. %

With reference to Fig.~\ref{fig:frame}, the NWU (X North, Y West, Z Up) is used as the inertial frame $\mathcal{F}_{I}$ , while the FLU (X Forward, Y Left and Z Up) is selected for the body-fixed frame $\mathcal{F}_{B}$. We also define unit vectors $e_j \in \mathbb{R}^3$ to denote the direction of arm $j\in\{1,2,3,4\}$ in body frame. For example, $e_1 = [1/\sqrt{2}, 1/\sqrt{2}, 0]^T$ and $\ e_2 = [1/\sqrt{2}, -1/\sqrt{2}, 0]^T$. 

We make the following assumptions in this work.  
\begin{enumerate}
        \item Rotor drag, moment generated by propeller angular speeds, and friction of the prismatic joint on compliant arms are ignored.
        \item The motors are encircled by protective cages that retain the shape during collisions. 
        \item During impact, the robot is modeled as five distinct rigid bodies.  
		\item During impact, the rigid body of arm $j$ only has one degree of freedom (dof) along $e_j$. 
		\item In free flight, the robot remains a single rigid body.
		\item The contact only results in kinetic friction with the obstacle.
\end{enumerate}
\vspace{0pt}

Note that based on assumption (4), all rigid bodies share the same orientation and angular velocity, thus $\omega$ and $q$ are independent of index $j$. 

\subsection{Rigid MAV Modeling}	
Based on assumption (5) and experimental validation (Section~\ref{sec:results}), we model the robot as a single rigid body when in free flight with a total mass $m = m_b + 4m_a$. 
Thus, robot position, velocity and their derivatives share the same values (e.g., $p_b = p_1 = p_2 = p_3 = p_4$). For clarity, we drop the index in $p$ and $v$. Equations of motion are derived using the Newton-Euler formulation as

\begin{equation}\label{Eqn:freeflight}
		\begin{aligned}
		\dot{p}  &= v   \\
		m\dot{v}  &= f_T Re_z - m g e_Z   \\
		       \dot{R}  &= R\hat{\omega}   \\
			I\dot{\omega} + \omega \times I\omega   &= \tau =  \begin{bmatrix} \tau_{\phi}\\ \tau_{\theta}\\\tau_{\psi} \end{bmatrix}
		\end{aligned}
	\end{equation}
where operator $\times$ denotes cross product. The hat map $\hat{\cdot}: \mathbb{R}^3 \xrightarrow{} \mathsf{SO(3)} $ is defined so that if $a,b \in \mathbb{R}^3, a \times b = \hat{a}b$. The robot is in $\mathsf{X}$ configuration (Fig.~\ref{fig:frame}), hence $f_T$ and $M$ are  
\vspace{-3pt}
\begin{equation}\label{Eqn:f_m}
		\begin{bmatrix} f_T\\\tau_{\phi}\\ \tau_{\theta}\\\tau_{\psi}\end{bmatrix} = 
		\begin{bmatrix} 1&1&1&1\\
		                L^\star& -L^\star& -L^\star& L^\star\\
		                - L^\star& - L^\star &  L^\star &  L^\star\\
		                -c_\tau & c_\tau & -c_\tau & c_\tau
		                
		\end{bmatrix} 
		\begin{bmatrix} f_{T,1}\\ f_{T,2}\\ f_{T,3}\\ f_{T,4}\end{bmatrix}\enspace,
	\end{equation}
where $L^\star = L/\sqrt{2}$, and $c_\tau$ is the moment coefficient. 

\subsection{Compliant MAV Modeling}
According to assumption (3), the system includes multiple rigid bodies under contact. In order to eliminate internal reaction forces, we study the dynamics modeling using the Euler-Lagrange method. Equations of motion build on top of prior work~\cite{Castillo}, with the additional inclusion of compliance and impact terms. We use Euler angles to represent the orientation of the robot as $\eta = [\phi, \theta, \psi]^T$ for roll, pitch and yaw angles, respectively. We also define a vector of the length of compliant arms $l = [l_1, l_2, l_3, l_4]^T \in \mathbb{R}^4$. Note the derivative $\dot{l} = [\dot{l}_1, \dot{l}_2,\dot{l}_3,\dot{l}_4]$ represents velocities of arm length changes in body frame. Similarly, we drop subscript $b$ for position $p$ and velocity $v$ of the main body for clarity of presentation. Then we can write the generalized coordinates in a vector $\mathcal{Q} \in \mathbb{R}^{10}$,
\vspace{-4pt}
\begin{equation}\label{Eqn:q}
\mathcal{Q} = [p, \eta, l]^T = [x,y,z, \phi, \theta, \psi, l_1, l_2, l_3, l_4]^T
\enspace.
\end{equation}
The Lagrangian $\mathcal{L}$ is calculated by the difference between kinetic $\mathcal{T}$ and potential $\mathcal{U}$ energy, that is
\vspace{-4pt}
\begin{equation}
\mathcal{L}(\mathcal{Q}, \dot{\mathcal{Q}}) = \mathcal{T} - \mathcal{U}\enspace.
\vspace{-2pt}
\end{equation}
The kinetic energy of the system includes both translational $\mathcal{T}_T$ and rotational $\mathcal{T}_R$ motion of all rigid bodies. We have
\begin{equation}\label{Eqn:kinetic}
		\begin{aligned}
		p_j  &= p + l_jRe_j  \\
		v_j  &= v + \dot{l}_jRe_j + l_jR\hat{\omega}e_j     \\
		\mathcal{T}_T &= \frac{1}{2}m_b v^Tv + \frac{m_a}{2}\sum_{i=1}^{4}v_i^Tv_i\\
		\mathcal{T}_R &= \frac{1}{2}\omega^TI\omega\\
		\mathcal{T} &= \mathcal{T}_T + \mathcal{T}_R \enspace.
		\end{aligned}
	\end{equation}
Note that the moment of inertia matrix $I$ of compliant MAVs is dependent on $l$ (estimated via Hall-effect sensor readings).

There are two core contributions to the potential energy $\mathcal{U}$ in the system: masses contribute gravitational potential energy, and elastic springs add stored energy. We adopt the Kelvin-Voigt (KV) model for compliant arms as in~\cite{cherouvim2009control, vasilopoulos2014compliant}. Thus, we can write the arm compressing forces as 
\vspace{-2pt}
\begin{equation}
f_L = k_l \delta_{L} + b_l \dot{\delta}_{L}
\vspace{-4pt}
\end{equation}
where $\delta_L$ denotes the arm length changes in body frame, and $k_l$ and $b_l$ stand for the spring and damping coefficients, respectively. Then the potential energy $\mathcal{U}$ can be found as
\begin{equation}\label{Eqn:potential}
		\begin{aligned}
		z_j  &=p_j^Te_Z \\
		\mathcal{U} &= m_b g z + m_a g \sum_{i=1}^{4} z_i + \frac{1}{2}k_l \sum_{i=1}^{4} (L-l_i)^2\enspace.
		\end{aligned}
\end{equation}

We use $F_{\text{ext}}\in \mathbb{R}^{10}$ to denote the generalized external force. Then, we can write the Euler-Lagrange equations as 
\begin{equation}\label{Eqn:eula}
		\begin{aligned}
		\frac{\text{d}}{\text{d}t}\Bigg(\frac{\partial \mathcal{L}}{\partial \dot{\mathcal{Q}}}\Bigg) - \Bigg(\frac{\partial \mathcal{L}}{\partial \mathcal{Q}}\Bigg)   &= F_{\text{ext}} = \frac{\partial}{\partial \dot{\mathcal{Q}}} \mathcal{P}_{\text{ext}} \\
		\end{aligned}
\end{equation}	
where $\mathcal{P}_{\text{ext}}$ is the power generated by external forces. 

We can split $\mathcal{P}_{\text{ext}}$ into three parts: $\mathcal{P}_{\text{rigid}}$, $\mathcal{P}_{\text{arm}}$ and $\mathcal{P}_{\text{contact}}$. As in rigid MAVs~\cite{Castillo}, $\mathcal{P}_{\text{rigid}}$ consists of the force and torque generated by motors,
\vspace{-2pt}
\begin{equation}\label{p_rigid}
\mathcal{P}_{\text{rigid}} = [0,0,f_T]\cdot v + \tau^T\omega
\vspace{-4pt}
\end{equation}
where $f_T$ and $\tau$ can be calculated similar to \eqref{Eqn:f_m} as

\begin{equation}\label{Eqn:f_m_2}
		\begin{bmatrix} f_T\\\tau_{\phi}\\ \tau_{\theta}\\\tau_{\psi}\end{bmatrix} = 
		\begin{bmatrix} 1&1&1&1\\
		                l_1^\star& -l_2^\star& -l_3^\star& l_4^\star\\
		                - l_1^\star& - l_2^\star &  l_3^\star &  l_4^\star\\
		                -c_\tau & c_\tau & -c_\tau & c_\tau
		\end{bmatrix} 
		\begin{bmatrix} f_{T,1}\\ f_{T,2}\\ f_{T,3}\\ f_{T,4}\end{bmatrix}
	\end{equation}
with $l_j^\star = l_j/\sqrt{2}$, and $c_\tau$ the same as in \eqref{Eqn:f_m}. 
Using the Rayleigh dissipation function, we can write $\mathcal{P}_{\text{arm}}$ as 
\begin{equation}\label{p_arm}
\mathcal{P}_{\text{arm}} = \frac{1}{2} b_l \sum_{i=1}^{4} \dot{l}_i^2\enspace.
\end{equation}
Note that we ignore the friction of the prismatic joints, therefore $\mathcal{P}_{\text{arm}}$ only includes springs along compliant arms. 

Lastly, we study the power of contact force $\mathcal{P}_{\text{contact}}$. 
The obstacle is assumed to be a vertical wall perpendicular to the $e_X$. Details about obstacles and generating contact forces will be elaborated in the next subsection. Here we summarize that there are $f_n, f_f \in \mathbb{R}^3$ in inertial frame, which represent  normal and frictional forces generated by the contact (Fig.~\ref{fig:frame}). We also have the velocity of the contact point $v_c$ in inertial frame. Then we can calculate $\mathcal{P}_{\text{contact}}$ as
\begin{equation}\label{p_contact}
\begin{aligned}
\mathcal{P}_{\text{contact}} &= f_f \cdot (v_c^Te_Z) - f_n \cdot (v_c^T e_X)\\
 \mathcal{P}_{\text{ext}}    &=  \mathcal{P}_{\text{rigid}} + \mathcal{P}_{\text{arm}} + \mathcal{P}_{\text{contact}}
 \end{aligned}
\end{equation}

\subsection{Impact Modeling}
\begin{figure}[!t]
\vspace{6pt}
	\centering
	\includegraphics[width=0.85\linewidth]{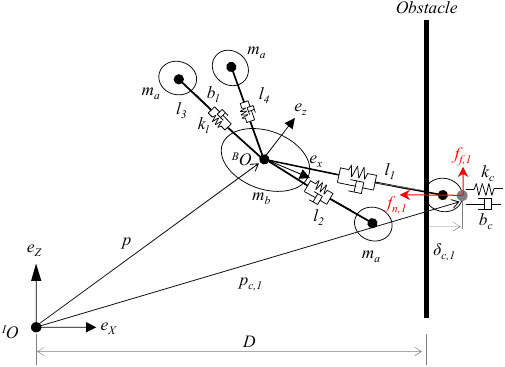}
	\caption{Contact geometry elements for a sample impact scenario with arm $j=1$. The obstacle is perpendicular to the $e_X$ at a known distance $D$. The contact point $p_{c,1}$ lies on the cage circle along arm, and collinear with $e_X$. The normal force $f_{n,1}$ and frictional force $f_{f,1}$ generated by contact in inertial frame are plotted in red. }
	\label{fig:model}
	\vspace{-18pt}
\end{figure}

In this paper, we consider the obstacle to be a vertical wall perpendicular to the $e_X$ with a known distance $D$, similar to~\cite{chui2016dynamics}. This selection simplifies the problem, however, our method can be extended to other obstacles by taking geometric constraints into account~\cite{dicker2018recovery}. 
From assumption (2), the cages retain their shape during impact, thus the contact point $p_{c,j}$ lies on the cage circle along arm $j$, and collinear with $e_X$. The contact geometric elements of the robot with one arm in contact are shown in Fig.~\ref{fig:model}. 


To study the contact model, we need to identify the location and velocity of the contact point $p_{c,j}$ along arm $j$ in inertial frame. The protective cage has a constant radius $r_0$, thus the contact point can be found by projecting the vector $r_0e_j$ in body frame onto the inertial $X$ axis. Then,
\begin{equation}\label{Eqn:contact_point}
		\begin{aligned}
        p_{c,j} &= p_j + \Big((r_0Re_j)^Te_X\Big)e_X\\
        v_{c,j} &= v_j + r_0\Big(e_j^T \Big(\frac{\text{d}}{\text{d}t}R^T\Big)e_X\Big)e_X\\
        \frac{\text{d}}{\text{d}t}R^T &= \frac{\text{d}}{\text{d}t}R^{-1} = -R^TR\hat{\omega}R^T
		\end{aligned}
	\end{equation}
where $p_j$ and $v_j$ can be found via \eqref{Eqn:kinetic}. The normal and tangential components of $v_{c,j}$ in the inertial frame are

\begin{equation}\label{Eqn:components}
		\begin{aligned}
        v_{c,j}^n &= (v_{c,j}^Te_X)e_X\enspace,\\
        v_{c,j}^t &= (v_{c,j}^Te_Z)e_Z\enspace.
		\end{aligned}
	\end{equation}

Based on the HC model~\cite{hunt1975coefficient}, the normal force $f_{n,j}$ and frictional force $f_{f,j}$ generated by contact on arm $j$ in inertial frame are modeled as


\begin{equation}\label{Eqn:contact}
		\begin{aligned}
    	f_{n,j} &= 	\begin{cases}
            \mathbf{0} &\text{for $\delta_{L,j} < 0$}\\
            k_c \delta_{L,j}^n - b_c \delta_{L,j}^n \dot{\delta}_{L,j} &\text{for $\delta_{L,j} 	\geq 0$}\end{cases}\\
    	f_{f,j} &=  -\mu f_{n,j} \frac{v_{c,j}^t}{||v_{c,j}^t||}
		\end{aligned}
	\end{equation}
where $\delta_{L,j} = p_{c,j}^Te_X - D $, and $ \dot{\delta}_{L,j} = v_{c,j}^n$; $\mu$ denotes is the kinetic friction coefficient. Using \eqref{p_contact}, we can calculate the power of contact force if only arm $j$ collides with the wall.  

For the two arms case, we can repeat \eqref{Eqn:contact_point} to \eqref{Eqn:contact} and calculate the velocities of the contact point $v_j$, as well as normal and frictional forces, respectively. For instance, if arms 1 and 2 are in contact, we can rewrite \eqref{p_contact} as
\begin{equation}\label{p_contact_new}
\begin{aligned}
\mathcal{P}_{\text{contact, j}} &= f_{f,j} \cdot (v_{c,j}^Te_Z) - f_{n,j} \cdot (v_{c,j}^T e_X)\\
\mathcal{P}_{\text{contact}} &= \mathcal{P}_{\text{contact, 1}} + \mathcal{P}_{\text{contact, 2}}\\
 \end{aligned}
\vspace{-6pt}
\end{equation}

\section{Motion Control and Collision Handling}\label{sec:impact}

In order to stabilize from high-speed and large-angle collisions, the robot controller must be able to track aggressive trajectories with large Euler angles. Our tracking control is based on~\cite{liu2021arq}. The position controller takes advantages of geometric constraints for nonlinear tracking as in~\cite{lee2010geometric, mellinger2011minimum, thomas2016aggressive}. The controller reads estimated current states ($p,v, \psi$) and desired states ($p_\text{des},v_\text{des},\ a_\text{des}, \psi_\text{des}$), and outputs the desired total thrust $f_\text{T, des}\in \mathbb{R}$ and desired attitude $R_\text{des} \in \mathsf{SO(3)}$.\footnote{~Although the project uses a Quaternion-based attitude control method~\cite{brescianini2013nonlinear}, other nonlinear attitude controllers can track the desired attitude from the position controller.} 

Collision handling to drive the robot to a safe position post collisions is based on~\cite{liu2021arq, tomic2017external}. The recovery setpoint $p_{r, \text{des}}$ the inertial frame can be found as 
\begin{equation}
    p_{r, \text{des}} = p_c - \delta f^0_c 
\end{equation}
where $\delta \in \mathbb{R}$ is a user-defined coefficient that can be tuned empirically. Details on how to fuse sensor readings to characterize the external impact are discussed in~\cite{liu2021arq}.



\section{RESULTS}\label{sec:results}

\begin{figure}[!h]
\vspace{4pt}
\centering
\begin{subfigure}{.45\linewidth}
  \centering
  \includegraphics[width=.95\linewidth]{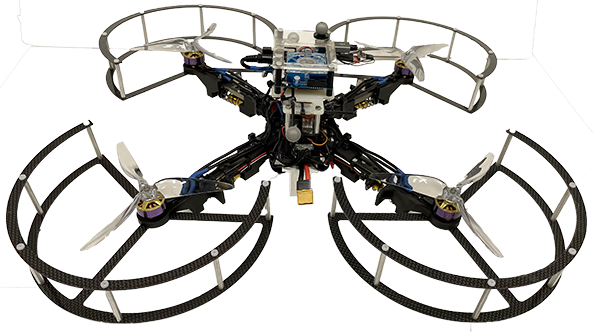}  
  \caption{ARQ}
\end{subfigure}
\centering
\begin{subfigure}{.45\linewidth}
  \centering
  \includegraphics[width=.95\linewidth]{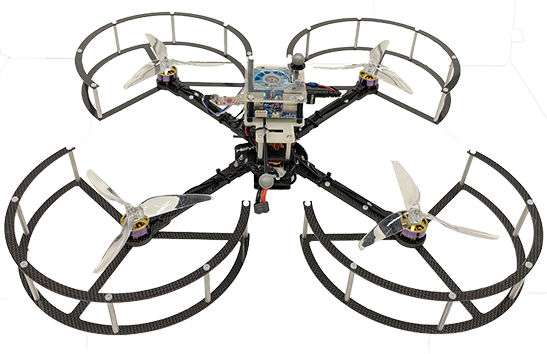}  
  \caption{Quad}
\end{subfigure}
\vspace{-5pt}
\caption{Prototypes of our (a) compliant and (b) rigid robots. 
}
\label{fig:hw_2}
\vspace{-16pt}
\end{figure}

To fully study the free flight performance and impact resilience of ARQ, we built another rigid robot termed herein as Quad. The rigid robot shares almost the same design with ARQ, except that rigid arms fabricated in lightweight carbon fiber sheets are used. Perspective views of both ARQ and Quad prototypes are shown in Fig.~\ref{fig:hw_2}. Key features of Quad are also listed in Table~\ref{tb:arq_values}. 

\subsection{Model Validation}
In this test, we fix both the rigid and compliant robots on a custom-made testbed with a linear slider 
that allows for horizontal-only (i.e. along the $e_X$ direction) motion (see supplemental video\;
As this part only focuses on the passive response to collisions, motors on both robots are not actuated during the tests. 
We examine collisions with two types of surfaces: rigid walls and soft mats. Motion capture at $200$\;Hz is utilized to measure positions and velocities of both the main body and arm. The distance $D$ of the obstacle is fixed and known before collisions ($D = 1.0$\;m). Ten repeated tests are conducted for each collision type for each robot.

In this study, we only focus on the short period of time during which robots are in contact with obstacles. Thus, we ignore the friction of the slider. Both robots are manually accelerated to velocities $\sim1.85$\;m/s right before collision with a zero contact angle $\theta = 0$. Note that the masses of both robots in this test also include the weight of the sliding bar. 
Several key parameters and their values are listed in Table~\ref{tb:model_value}. Similar to the related work~\cite{vasilopoulos2014compliant}, we set the order of the impact model $n=1.5$ and damping coefficient $b_c = 1.5\cdot c_a \cdot k_c$, where $c_a$ is usually between 0.01 - 0.5 depending on the materials and impact velocity.


\begin{table}[htb]
\vspace{-3pt}
\renewcommand{\arraystretch}{1.3}
\caption{Key Parameters of Model Validation Study}
\label{tb:model_value}
\vspace{-0pt}
\centering
\begin{tabular}{c c c c c}
\toprule
 $m_b$& $m_a$&  $m_\text{rigid}$ & $\mu$ \\
\midrule
 $1.882$\; kg  &  $0.2$\; kg &  $1.892$\; kg & $0.8$  \\
\midrule
\midrule
 $k_l$ & $b_l$ & $k_{c,\text{wall}}$ &  $k_{c,\text{mat}}$ \\
 \midrule
 $5\times 10 ^3 $\; N/m & $90$\; Ns/m &  $2\times 10 ^5 $\; N/m& $1.2\times 10^5 $\; N/m  \\
\bottomrule
\end{tabular}
\vspace{-6pt}
\end{table}

\begin{figure}[!h]
\vspace{2pt}
	\centering
	\includegraphics[width=0.99\linewidth, trim={0.5cm 5mm 0.5cm 1mm}]{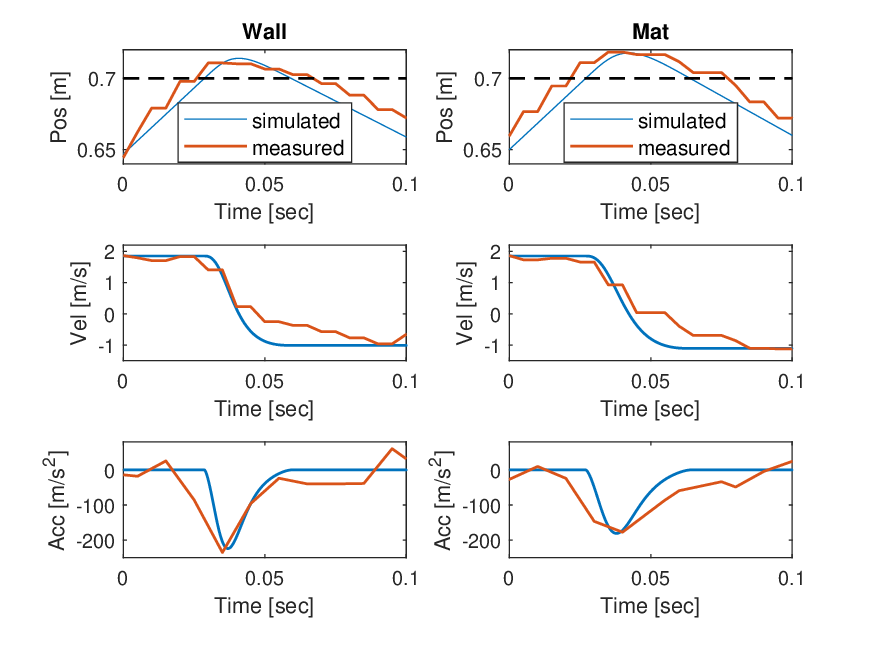}
	\vspace{-15pt}
	\caption{Positions, velocities and accelerations of the rigid robot (QUAD) in collisions with both rigid walls and soft mats.}
	\label{fig:impact_rigid_re}
	\vspace{-16pt}
\end{figure}

Results of the Quad colliding with rigid walls and soft mats are shown in Fig.~\ref{fig:impact_rigid_re}. Positions, velocities and accelerations are shown in blue and red curves to describe results in simulation and physical tests, respectively. The black dashed lines are used to denote the position when the robot is in contact with obstacles. Note that the collision positions are different in the mat tests due to the thickness of mats, however, they are re-aligned to match those in the rigid wall tests. Deformation is observed in both cases with $\delta_\text{wall} = 14$\;mm and $\delta_\text{mat} = 18$\;mm. We also observe very short contact time as $\text{dt}_\text{wall} = 0.031$\;sec and $\text{dt}_\text{mat} = 0.037$\;sec. 

Despite the discontinuity in the measured data due to the frequency limitation of the motion capture, simulated results still fit the physical experimental data well. 
We list the coefficient of restitution (COR) and maximum absolute accelerations $a_\text{max}$ of both simulated and physical tests in Table~\ref{tb:rigid_impact_co}. COR is computed as the ratio of the absolute post velocity over the absolute prior one ($\text{COR} = |v_{\text{after}}| / |v_{\text{before}}|$). Based on the results, we conclude that the adopted continuous impact models can describe collisions with various surfaces well. However, we observe large impact for the rigid robot, especially in collisions with rigid walls ($a_\text{max} = 235$\;m/s$^2$), which may damage sensitive electronic components. 

\begin{table}[htb]
\vspace{-1pt}
\centering
  \caption{Rigid and Compliant Robots Impact Study}
  \vspace{-4pt}
\begin{tabular}{ccccc}
\multicolumn{5}{c}{Quad}\\    
\toprule
  & \multicolumn{2}{c}{Simulation} & \multicolumn{2}{c}{Physical Test} \\
\cmidrule(lr){2-3}\cmidrule(lr){4-5} 
 & Wall & Mat  & Wall & Mat  \\ \midrule

COR & 0.547 & 0.598  & 0.526 & 0.603 \\ \midrule
$a_\text{max}\ [\text{m/s}^2]$& 224 & 168  & 235& 177  \\ 
\bottomrule

& & & & \\

\multicolumn{5}{c}{ARQ}\\    
\toprule
  & \multicolumn{2}{c}{Simulation} & \multicolumn{2}{c}{Physical Test} \\
\cmidrule(lr){2-3}\cmidrule(lr){4-5} 
 & Wall & Mat  & Wall & Mat  \\ \midrule
COR& 0.444 & 0.498 & 0.451 &  0.504\\ \midrule
$a_\text{max}\ [\text{m/s}^2]$& 144 & 137  & 146 & 142  \\ 
\bottomrule

\end{tabular}%
\label{tb:rigid_impact_co}
\vspace{-4pt}
\end{table}

\begin{figure}[!t]
\vspace{6pt}
	\centering
	\includegraphics[width=0.9\linewidth, trim={0.5cm 8mm 0.5cm 0.9cm},clip]{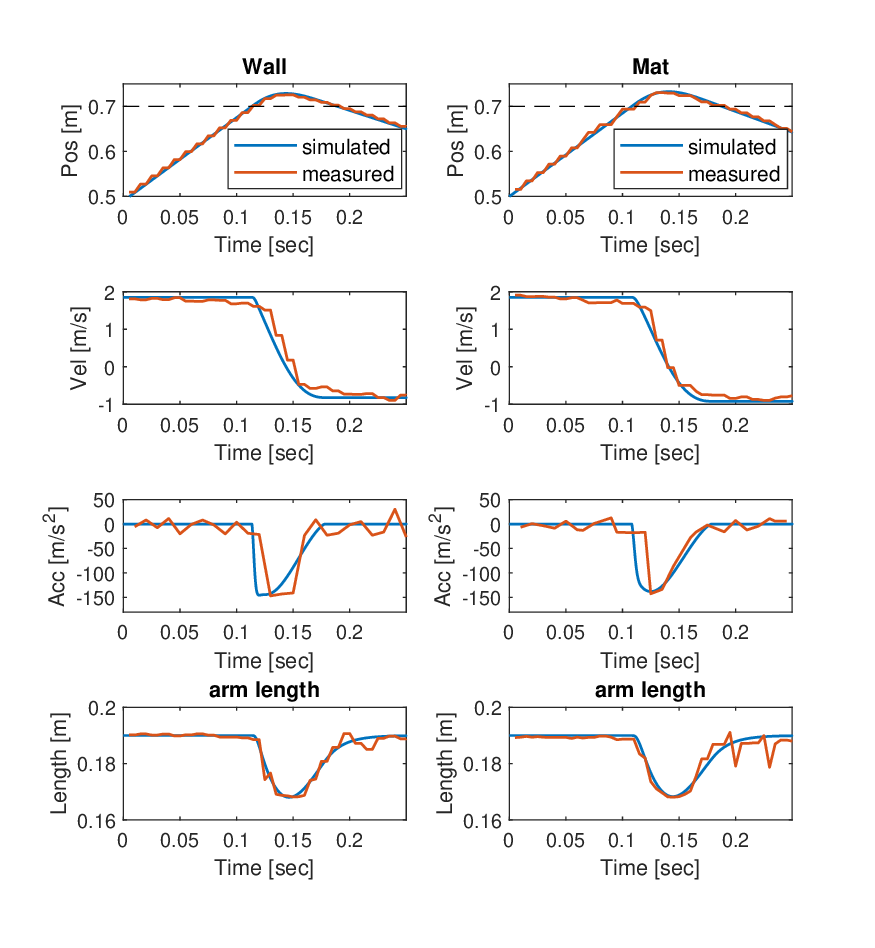}
	\vspace{-5pt}
	\caption{The states of the compliant robot's (ARQ) main body, as well as arm length, in collisions with both rigid walls and soft mats.}
	\label{fig:impact_arq_re}
	\vspace{-18pt}
\end{figure}

Results from ARQ impact experiments with rigid walls and soft mats in both simulation and physical tests are shown in Fig.~\ref{fig:impact_arq_re}. Besides positions, velocities and accelerations of the main body, the length of the compliant arms are also included in the figure. 
Due to the compliant arms, larger deformation is observed for collisions with both surfaces ($\delta_\text{wall} = 29$\;mm, $\delta_\text{mat} = 33$\;mm). A longer contact time is also observed for compliant robot collisions as $\text{dt}_\text{wall} = 0.079$\;sec and $\text{dt}_\text{mat} = 0.083$\;sec. Despite noise in measured curves, simulated results generally fit those from physical tests well, including the compliant arm length changes. Results validate the proposed model for compliant MAVs under contact.   

The COR and maximum absolute accelerations of the compliant robot are also listed in Table~\ref{tb:rigid_impact_co}. Simulated results again match the physical ones well. Compared to Quad, the compliant robot has smaller COR and $a_\text{max}$, owing to the impact reduction afforded by the compliant arms. Moreover, smaller differences on both COR and $a_\text{max}$ between different surfaces are observed, due to the presence of compliant arms.  

We notice that maximum absolute accelerations $a_\text{max}$ are dependent on the contact velocity. We conduct additional simulated tests to record the maximum absolute accelerations for both rigid and compliant robots in collision with rigid walls. Parameters in the simulation are the same to those in the study above, with one exception that we only include the masses of the robots (without the weight of the sliding bar).  

\begin{figure}[!h]
\vspace{-9pt}
	\centering
	\includegraphics[width=0.55\linewidth]{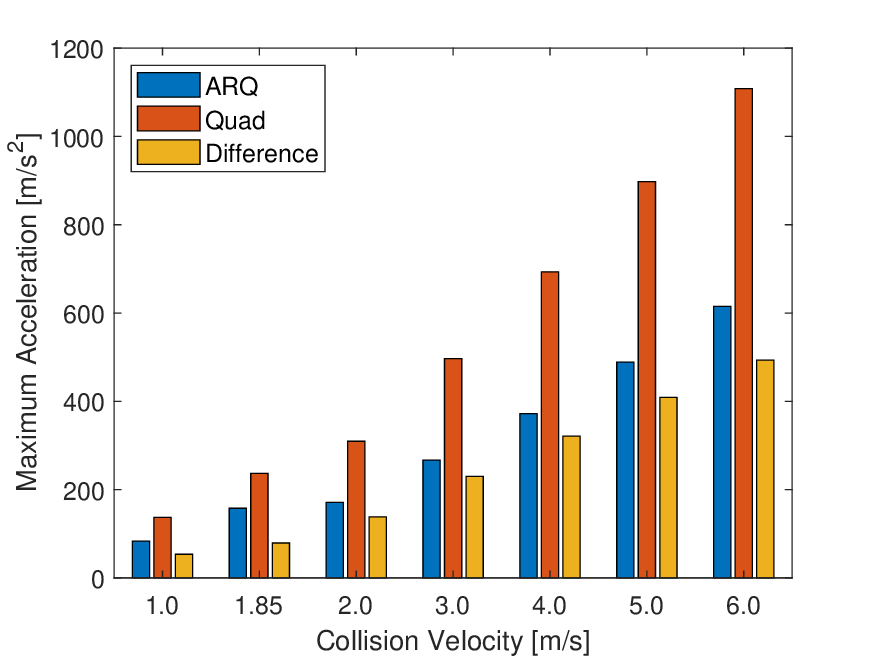}
	\vspace{-3pt}
	\caption{Maximum absolute acceleration in extended simulation tests.}
	\label{fig:acc_impact}
	\vspace{-12pt}
\end{figure}

Figure~\ref{fig:acc_impact} visualizes the simulated results with collision speeds ranging from $1$\;m/s to $6$\;m/s. Blue bars denote the maximum absolute accelerations of the compliant robot while the red bars represent those of the rigid MAV. We also use yellow bars to visualize differences between compliant and rigid robots. Simulated results support our claim that the compliant arms play a big role in reducing the impact to the main body under high-speed contact, as we see a difference of $\text{d} a_\text{max} = 493$\;m/s$^2$ or $50.26$\;G at a speed of $6$\;m/s. 

\subsection{Impact Resilience}


We use a wooden wall as the obstacle (see Fig.~\ref{fig:arq_35}), which has hardness close to rigid walls in the previous test. Both compliant and rigid robots are running the nonlinear tracking controller, and following trajectories generated offline before colliding with two arms in contact. The position of the wall is unknown to the robots before collisions; ground truth is measured only for reference ($D = 2.75$\;m). The compliant robot uses the detection method with Hall effect sensors while the rigid robot utilizes the accelerometer to report collisions. Both robots adopt the same recovery method with a constant distance $L_c = 0.1$\;m. Note that the body-fixed frame $\mathcal{F}_{B}$ is aligned with the inertial frame $\mathcal{F}_{I}$ ($R = I_3$) before the robot starts collision tests, therefore both robots collide with the obstacle with two arms in contact (Fig.~\ref{fig:arq_35}). 

Over 130 collision tests are conducted for the two robots. Two velocities ($2.0$\;m/s and $3.5$\;m/s) are studied with a zero contact angle for both compliant and rigid robots. We also examine collisions at a constant velocity $2.0$ m/s but with various pitch angles ($-45^\circ, \pm 30^\circ, \pm 15^\circ$). Ten consecutive experimental trials are implemented for each case for both robots (e.g., Quad at $2$ m/s with angle $0^\circ$).

\begin{figure}[!t]
\vspace{10pt}
\centering
\begin{subfigure}{.45\linewidth}
  \centering
  \includegraphics[trim={1cm 0.8cm 1cm 1cm}, width=.98\linewidth]{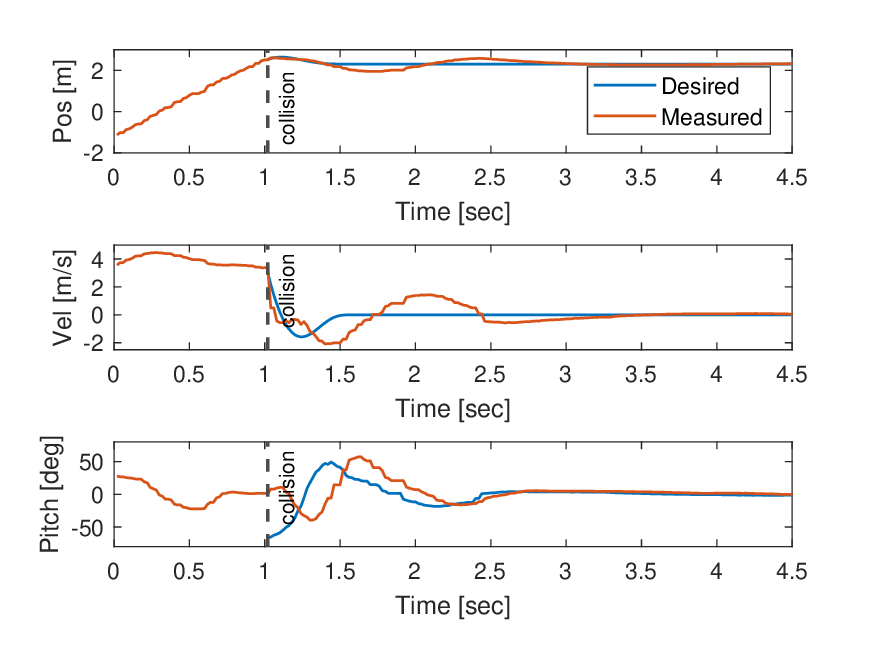}  
  \caption{States}
  \label{fig:arq_35_states}
\end{subfigure}
\centering
\begin{subfigure}{.45\linewidth}
  \centering
  \includegraphics[trim={1cm 0.5cm 1cm 1cm}, width=.98\linewidth]{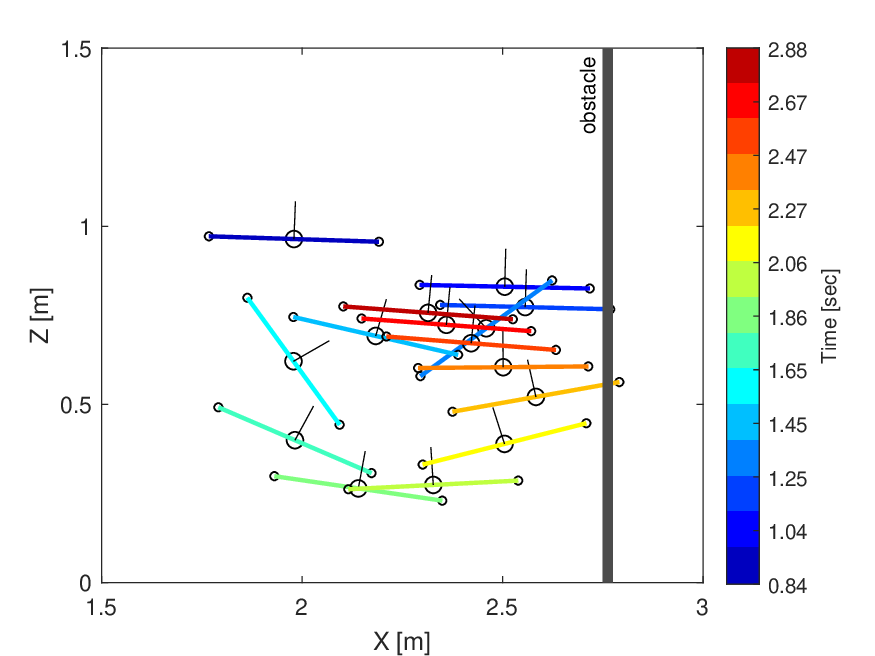}  
  \caption{Orientation}
  \label{fig:arq_35_traj}
\end{subfigure}
\vspace{-3pt}
\caption{(a) States and (b) orientation of the compliant robot in a collision at the speed of $3.5$\;m/s.}
\label{fig:arq_35_contact}
\vspace{-12pt}
\end{figure}

Results show that ARQ can survive collisions and sustain flight at a speed of $3.5$ m/s with zero contact angles, at 100\% success rates. Composite images of a sample test can be found in Fig.~\ref{fig:arq_35}. States (position $p_x$, velocity $v_x$ and pitch angles $\theta$) of the main body are shown in Fig.~\ref{fig:arq_35_states}. 
We also visualize the projection of the robot's orientation onto the $e_X - e_Z$ plane (Fig.~\ref{fig:arq_35_traj}), where short black lines denote the $e_z$ axis. Black circles in different sizes represent the main body (big) or arms (small). The color scale denotes the temporal duration of the collision recovery process. Note that we use a constant arm projected length $L^\star = L/\sqrt{2}$ to visualize orientation in Fig.~\ref{fig:arq_35_states}. Results also show that the recovery method outlined in Section~\ref{sec:impact} can generate pitch angle trajectories similar to but more aggressive than~\cite{dicker2018recovery} (e.g., $-66^\circ, 49^\circ$). Taking advantage of the nonlinear tracking controller, ARQ can follow the aggressive recovery trajectory and stabilize itself rapidly, before hovering stably.

\begin{figure}[!t]
\vspace{0pt}
	\centering
	\includegraphics[width=0.8\linewidth,trim={0 0.1cm 0 0.25cm},clip]{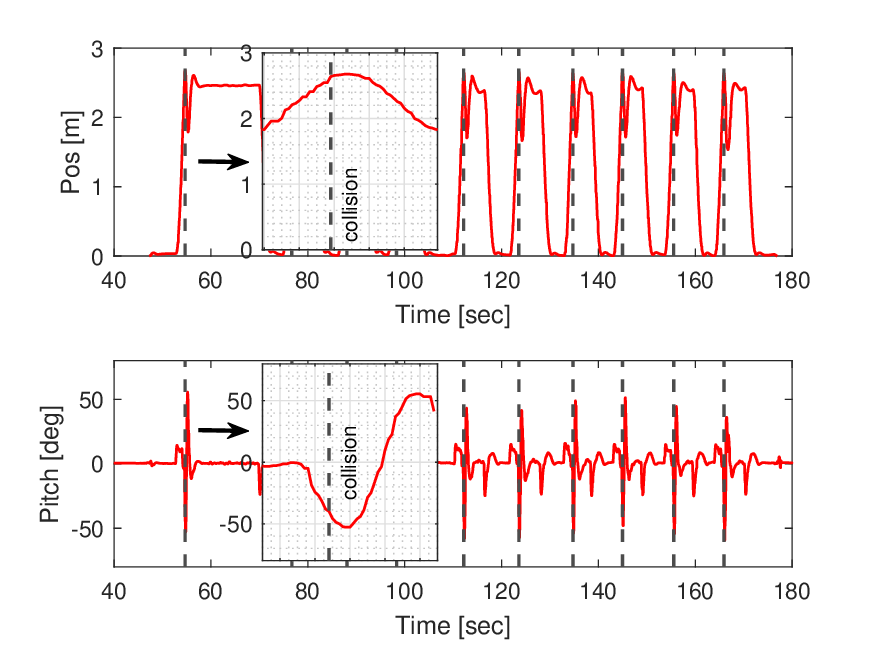}
	\caption{Repeated collision tests of the compliant robot with a contact angle of $-45^\circ$. The boxes show detailed (zoomed-in) portions of states (within $1$\;sec) for the first collision.}
	\label{fig:arq_n45}
	\vspace{-12pt}
\end{figure}


\begin{table*}[!t]
\vspace{3pt}
  \caption{Comparisons in Step Response and Planar Circle Tracking Tests}
\makebox[\textwidth][c]{
    \begin{tabular}{cccccccccc}
\multicolumn{10}{c}{Step Response}\\    
\toprule
  & \multicolumn{3}{c}{Quad} & \multicolumn{3}{c}{ARQ pre-col} & \multicolumn{3}{c}{ARQ post-col}\\
\cmidrule(lr){2-4}\cmidrule(lr){5-7} \cmidrule(lr){8-10}
 & $p_x$ & $p_y$  & $p_z$ & $p_x$ & $p_y$  & $p_z$ & $p_x$ & $p_y$& $p_z$ \\ \midrule

Rise time [sec]& 0.649 & 0.662  & 0.625 & 0.760 & 0.782 & 0.627 & 0.767 & 0.779 & 0.628 \\ \midrule
MSE [m]& 0.090 & 0.094  & 0.059& 0.098 & 0.107  & 0.064 & 0.108 & 0.108 & 0.062 \\ 
\bottomrule
\end{tabular}%
}
\bigskip

\makebox[\textwidth][c]{
    \begin{tabular}{cccccccccccccc}
    
    \multicolumn{14}{c}{Planar Circle Tracking}\\   
\toprule

   &   & \multicolumn{4}{c}{Quad} & \multicolumn{4}{c}{ARQ pre-col} & \multicolumn{4}{c}{ARQ post-col}\\
\cmidrule(lr){3-6}\cmidrule(lr){7-10} \cmidrule(lr){11-14}
& & $p_x$ & $p_y$  & $v_x$ & $v_y$ & $p_x$ & $p_y$  & $v_x$ & $v_y$ &  $p_x$ & $p_y$  & $v_x$ & $v_y$ \\ 
\cmidrule(lr){3-4} \cmidrule(lr){5-6} \cmidrule(lr){7-8} \cmidrule(lr){9-10} \cmidrule(lr){11-12} \cmidrule(lr){13-14}
   &   & \multicolumn{2}{c}{[m]}  & \multicolumn{2}{c}{[m/s]} & \multicolumn{2}{c}{[m]}  & \multicolumn{2}{c}{[m/s]}  & \multicolumn{2}{c}{[m]}  & \multicolumn{2}{c}{[m/s]}\\

\midrule

MSE & Slow  & 0.88 &  3.80  & 1.20 & 23.90 & 1.10 & 6.10  & 1.80 & 23.70 & 2.00 & 6.20 & 1.80 & 24.40\\ 

\cmidrule(lr){2-14}

$\times 10^{-3}$& Fast & 9.70 & 13.80  & 46.10 & 100.80 & 15.90 & 17.00  & 74.50 & 99.50 &  22.20 & 19.90  & 95.40 & 102.80 \\

\bottomrule
\end{tabular}%
}
\label{tab:freflight}%
\vspace{-12pt}
\end{table*}%

Compliant arms are also observed to significantly contribute to stabilization in high-speed collisions by reducing impact and elongating the contact time. In comparison, when the rigid robot collides at the same speed, it fails to stabilize and sustain flight in all trials. Owing to the compliant airframe, ARQ has 100\% success rates for both high-speed and large-angle collisions. The rigid robot has fairly good performance within the pitch range $-30^\circ$ to $30^\circ$ at a speed of $2$ m/s, however, the success rate drops when the pitch reaches $-45^\circ$. We present the position $p_x$ and pitch $\theta$ values of ten consecutive recovery trials of ARQ with pitch angles $-45^\circ$ in Fig.~\ref{fig:arq_n45}. The robot sustains flight and goes to the initial point $[0,0,1]^T$ after recovering from large-angle collisions. Then, the robot repeats a collision test without landing. In all, the 100\% success rates for stabilizing high-speed and large-angle collisions significantly promote the confidence in deploying ARQ in unknown challenging environments.

\begin{figure}[!t]
\vspace{9pt}
\centering
\begin{subfigure}{.45\linewidth}
  \centering
  \includegraphics[trim={1cm 0.8cm 1cm 1cm}, width=.98\linewidth]{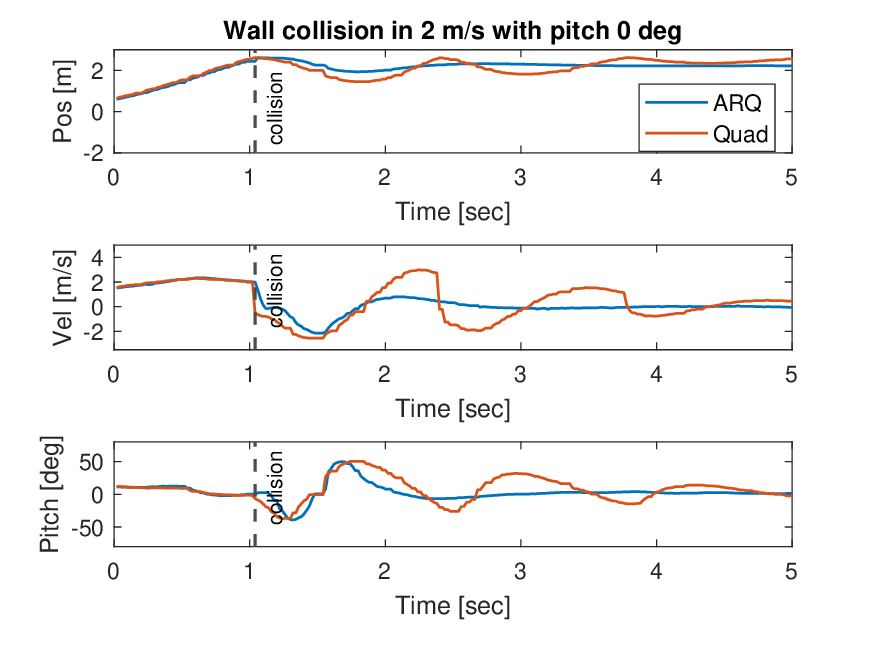} 
  \vspace{-12pt}
  \caption{States}
  \label{fig:2ms_two}
\end{subfigure}
\vspace{-3pt}
\centering
\begin{subfigure}{.45\linewidth}
  \centering
  \includegraphics[trim={1cm 0.5cm 1cm 1cm}, width=.98\linewidth]{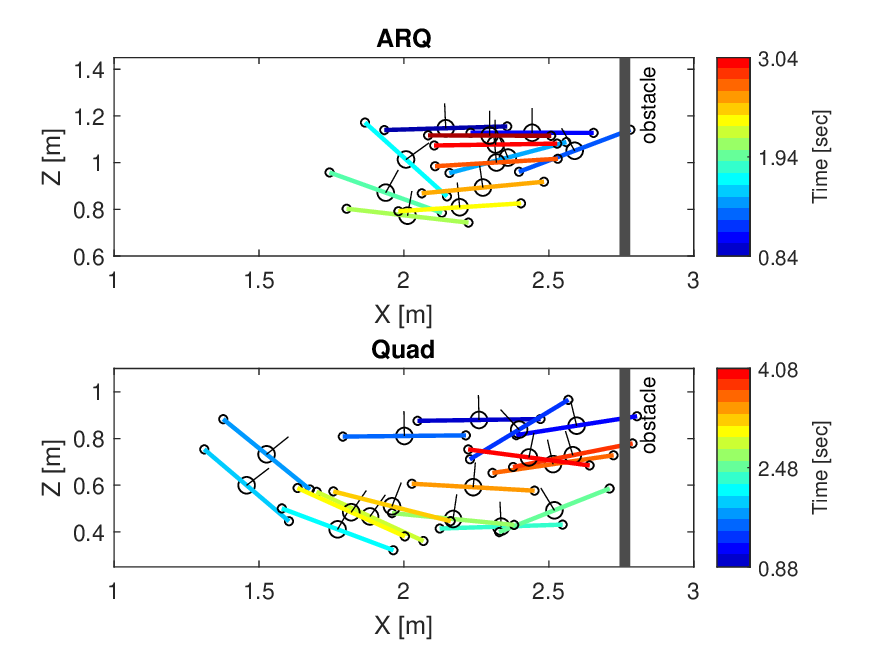}  
  \vspace{-12pt}
  \caption{Orientation}
  \label{fig:2ms_two_att}
\end{subfigure}
\caption{Sample states and orientation for the compliant and rigid aerial robots colliding at a speed of $2.0$ m/s with zero contact angle.}
\label{fig:2ms_two_overal}
\vspace{-18pt}
\end{figure}

Figure~\ref{fig:2ms_two} shows measured states (position $p_x$, velocity $v_x$, and pitch $\theta$) of two robots in a sample collision test at a speed of $2$ m/s with a $0^\circ$ pitch angle. 
When a collision happens, the velocity of ARQ declines less sharply, thanks to the impact reduction offered by the compliant airframe. The reduced impact and elongated contact time enable the compliant robot to stabilize itself rapidly, while oscillations occur to the rigid robot, resulting in a longer settling time. We also show the 2D orientation of both compliant and rigid robots after collision in Fig.~\ref{fig:2ms_two_att}. It can be seen that the rigid robot swings back farther compared to ARQ due to larger impact and shorter contact time. Both robots track large pitch angles to recover; yet, ARQ stabilizes much faster than Quad.


\subsection{Free Flight}

This experiment studies the free flight performance of ARQ before or after collisions, in comparison with the rigid robot Quad. Five repeated trials are recorded for each test. Note that the compliant robot is studied before (pre-col) and after (post-col) collisions separately. ARQ undergoes free falls at $1$\;m before the post-collision free flight performance study (see supplemental video). In the step response test, both robots hover at the point $[0,0,1]^T$ before the planner sends discrete setpoints $[0,0,2]^T$, $ [1,0,2]^T$ and $[1,1,2]^T$ at $5$\;sec intervals.  
Two planar circle trajectories are generated for the second test with periods of $2\pi$ and $\pi$\;sec, respectively. Both trajectories have radius of $1$ m, and start at $[1,0,1]^T$. (Note that positions in both tests are expressed in meters.) 


We calculate the rise time and mean squared errors (MSE), which are listed in Table~\ref{tab:freflight}. Herein the rise time is measured as the time the response takes to rise from 10\% to 90\% with respect to the attained steady-state value. Compared to the rigid robot, ARQ is observed to have slightly worse but in general very close performance for the step response test because of its increased body weight, in terms of both rise time and mean squared errors. 
Meanwhile, pre- and post-collision step response for ARQ closely matches each other. 


Similar observations can be made for the planar circle tracking test. Positions and velocities of both robots are tracked well for the slow circle, despite tracking errors caused by discontinuity of desired velocities. 
Compared to the rigid robot, ARQ has slightly worse but in general very close performance in the step response test, in terms of both rise time and mean squared errors.
However, both rigid and compliant robots are observed to have a minor difference in $x, y$ axis, resulting from the asymmetric hardware design.  


\section{CONCLUSIONS}

This paper presents new findings for an impact-resilient MAV equipped with passive springs within its compliant arms to reduce impact. Specifically, the paper contributes dynamic modeling to understand the effect of compliance. The impact resilience is reinforced to stabilize following high-speed and large-angle collisions. Furthermore, we perform detailed analysis and include comprehensive comparisons with a rigid MAV to validate the efficacy of our compliant robot. 
Physical collision tests are conducted for two robots to thoroughly examine impact resilience. Results show that the compliant ARQ robot can stabilize from high-speed and large-angle collisions rapidly and sustain post-impact flight. Free flight tests show that the compliant robot has very close tracking performance compared to its rigid counterpart.
This work introduced several interesting directions for further research. 
First, the impact modeling conducted herein only considers vertical walls as obstacles to recover from following a collision. 
Future work includes precise compliant MAV modeling under impact with diverse obstacles. Second, the project relies on motion capture for state estimation, which limits application to real-world environments. In future work, we plan to integrate onboard visual perception to the compliant robot and research impact resilience for aggressive autonomous flight in outdoor environments. 








\bibliographystyle{IEEEtran}
\bibliography{IEEEabrv, mybib}

\end{document}